\documentclass{article}

\usepackage[table]{xcolor}
\usepackage[utf8]{inputenc} 
\usepackage{hyperref}       
\usepackage{url}            
\usepackage{booktabs}       
\usepackage{amsfonts}       
\usepackage{microtype}      
\usepackage{xcolor}         
\usepackage{amsmath}        
\usepackage[linesnumbered, vlined, ruled]{algorithm2e} 
\usepackage{graphicx}
\usepackage{caption}
\usepackage{subcaption}
\usepackage{wrapfig}
\usepackage[square,numbers]{natbib}
\usepackage[margin=0.8in]{geometry}
\usepackage{authblk}

\title{Auxiliary Network: Scalable and agile online learning for dynamic system with inconsistently available inputs}

\author[1]{Rohit Agarwal}
\author[2]{Arif Ahmed Sekh}
\author[2]{Krishna Agarwal}
\author[3]{Dilip K. Prasad}
\affil[1]{Indian Institute of Technology (ISM),Dhanbad, India}
\affil[2]{Department of Physics and Technology,UiT The Arctic University of Norway}
\affil[3]{Department of Computer Science, UiT The Arctic University of Norway }
\date{}
\begin{document}

\maketitle

\begin{abstract}
  Streaming classification methods assume the number of input features is fixed and always received. But in many real-world scenarios demand is some input features are reliable while others are unreliable or inconsistent. In this paper, we propose a novel deep learning-based model called Auxiliary Network (Aux-Net), which is scalable and agile. It employs a weighted ensemble of classifiers to give a final outcome. The Aux-Net model is based on the hedging algorithm and online gradient descent. It employs a model of varying depth in an online setting using single pass learning. Aux-Net is a foundational work towards scalable neural network model for a dynamic complex environment requiring ad hoc or inconsistent input data. The efficacy of Aux-Net is shown on public dataset.
\end{abstract}


\section{Introduction}
\label{sec1}

\begin{figure}
\centering
    \begin{minipage}{0.45\textwidth}
    \includegraphics[width=\linewidth]{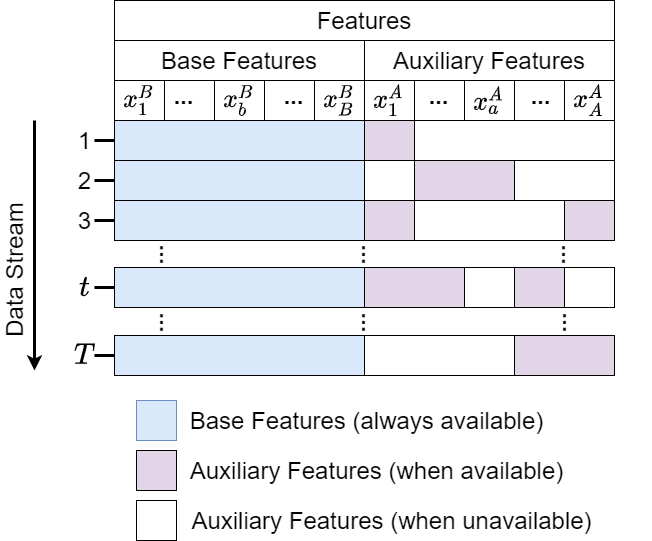}
    \caption{{Arrival of streaming data with all the base features and inconsistently available auxiliary features is demonstrated here.}}\label{fig1}
    \end{minipage}    \hfill
    \begin{minipage}{0.5\textwidth}
    \includegraphics[width=\linewidth]{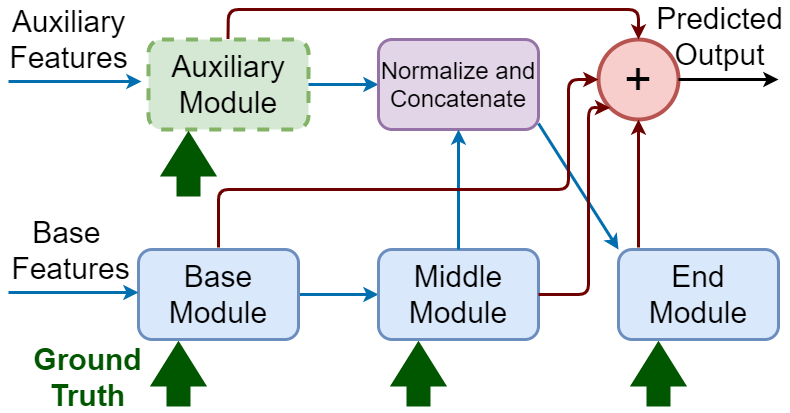}
    \caption{Block diagram of the auxiliary network. The green bold arrow represents the ground truth and the circle with a + sign calculates  a final prediction from the weighted output of each classifier. All the modules are the combination of one or more hidden layers where the auxiliary module is scalable i.e. the number of layers keeps changing with time and scalability needs of the application.}\label{fig2}
    \end{minipage}

\end{figure}

Machine learning architectures that support varying number of input features can be a game changer in many real life applications which deal with learning in dynamic complex environments. Examples include imparting intelligence to a node in ad hoc communication networks, a device in smart city environment, and an autonomous vehicle in complex driving environment. To model such dynamic environment of inconsistent and scalable nature with assumption some reliable data channels as base channels from on-board sensor array, which we refer to as base input features and denote as $\{x^{B}_1, \ldots, x^{B}_b, \ldots, x^{B}_B\}$. In addition, it may receive other information about the environment through auxiliary sensor arrays or communication channels. We call the corresponding input features as auxiliary input features, denote them by $\{x^{A}_1, \ldots, x^{A}_a, \ldots, x^{A}_A\}$.
Here, $x$ denotes input features, $B$ in superscript and subscript denotes base feature and the number of base features respectively. Similarly, $A$ in superscript and subscript denotes auxiliary feature and the number of auxiliary features respectively.
Due to the intermittent availability, only a subset of auxiliary features arrive along with the base features at any time instance $t$ as shown in Figure \ref{fig1}.
This problem can be approached in either  minimalist or maximalist approach. In the minimalist approach, all uncertain inputs are dropped and a single knowledge model is trained using only the base input features.
This knowledge model provides certain base accuracy, but does not utilize the additional information from auxiliary inputs.
The trade-off is the loss of opportunity for better performance.
In the maximalist approach, an ensemble of $2^A$ networks can be formed to cater for all possible combinations of availability of auxiliary features. Therefore the network with the smallest dimensionality caters to only the base features and the network with the largest dimensionality caters to all the base and auxiliary features, where we refer to the number of inputs to a network as its dimensionality. However, learning the knowledge model in such ensemble of networks is cumbersome, as explained next. Given $A_t$ inputs features at a time $t$, $2^{A_t}$ subsets of these features can be formed and therefore the network corresponding to each subset needs to be trained. This results into long training durations, further illustrated in supplementary. Another trade-off is that huge number of networks need to be maintained throughout. An ideal solution would be an agile and scalable network architecture that adapts itself to the availability of auxiliary inputs without needing to maintain or train multiple networks.

In this paper, we present a new paradigm of learning in the presence of inconsistently available auxiliary inputs, which we call auxiliary network (Aux-Net). The key-stone of Aux-net is the separation of learning corresponding to the auxiliary inputs and the base inputs into separate modules parallel to each other (see Figure \ref{fig2}). The base features are processed as a chunk in the base module while the auxiliary module contains one independent layer per auxiliary input in parallel with the other layers. Therefore, the dimensionality of the active network can be varied by simply freezing the portion corresponding to an unavailable auxiliary input. In this manner, the knowledge base of Aux-Net corresponds to the maximalist approach, where as the active knowledge model of suitable dimensionality can be invoked from it in an agile manner. Support for knowledge models of various dimensionalities makes our approach scalable. At the same time, agility of our framework and stability during dynamic operation is attributed to the special output weighing mechanism of the auxiliary block which dynamically spools the relative contribution of the auxiliary data depending on the availability of the auxiliary features and their influence on the final outcome.

We construct the initial framework on the basis of online deep learning (ODL) method \cite{sahoo2017online} and show our results on Italy power demand dataset \cite{dau2019ucr}. We observe robust, agile, and scalable performance of Aux-Net in situations as challenging as half of inputs being available 50\% of the time and when all except one inputs are intermittently available. We show that in the most challenging scenarios, Aux-Net performs quite close to ODL (trained using only the base features) even while supporting agility and in the more favorable scenarios, it performs better than ODL.

The outline of the paper is as follows. Related work is discussed in section \ref{sec2}. Aux-Net is presented in section \ref{sec3} and diverse numerical experiments and its discussion are presented in section \ref{sec4}. The paper is concluded in section \ref{sec5} and the broader impact of this work is presented in the last section.



\section{Related Works}
\label{sec2}

Many methods based on Bayesian theory \cite{seidl2009indexing}, k-nearest neighbour \cite{aggarwal2006framework}, support vector machines \cite{tsang2007simpler}, decision tree \cite{domingos2000mining}, fuzzy logic \cite{das2016ierspop, iyer2018pie} are proposed for streaming classification task. A brief study of all these techniques can be found in \cite{nguyen2015survey, gama2012survey}. Furthermore, some incremental learning approaches are also proposed \cite{polikar2001learn++, polikar2010learn++, muhlbaier2008learn, muhlbaier2007multiple, ditzler2010incremental}. Other deep learning approaches for online learning include \cite{das2019fernn, das2019muse, sahoo2017online, ashfahani2019autonomous}. A limitation of these techniques are that they assume that the dimension of the incoming data is fixed. Hou et al. \cite{hou2017one,hou2017learning} proposed machine learning approaches for dynamic environments. However, they assumed that the dimensionality of the inputs is constant in batches and therefore batch wise learning can be used.
Hou et al. \cite{hou2017one} further assume that there are multiple sets of features, where one entire set is either available or unavailable in a given batch.
Hou et al. \cite{hou2017learning} assumes that there is an overlapping period between two batches when all the inputs from previous and next batches are available, which allows in supporting soft transition across batches. These methods are indeed more scalable than approaches that assume fixed input dimensionality. Nonetheless, they cannot handle as challenging situations as depicted in Figure \ref{fig1} where no assumption is made on availability of auxiliary features in batches or sets. To the best of our knowledge, our work is a foundational work for problems in which the inputs may be inconsistently unavailable at any time instance. The only assumption of our work is that there is at least one base feature consistently available. Our work has a more general premise and the premises of all the above mentioned works can be considered as subsets of our premise. We note that our framework is inspired from the concept of hedge algorithm \cite{freund1995desicion} and online gradient descent (OGD) \cite{zinkevich2003online} used in deep neural network (DNN) as proposed in ODL \cite{sahoo2017online}, but ODL also assumes that all the features are consistently available.



\begin{figure}[t]
\centering
\begin{subfigure}{.6\textwidth}
  \centering
  \includegraphics[width=1\linewidth]{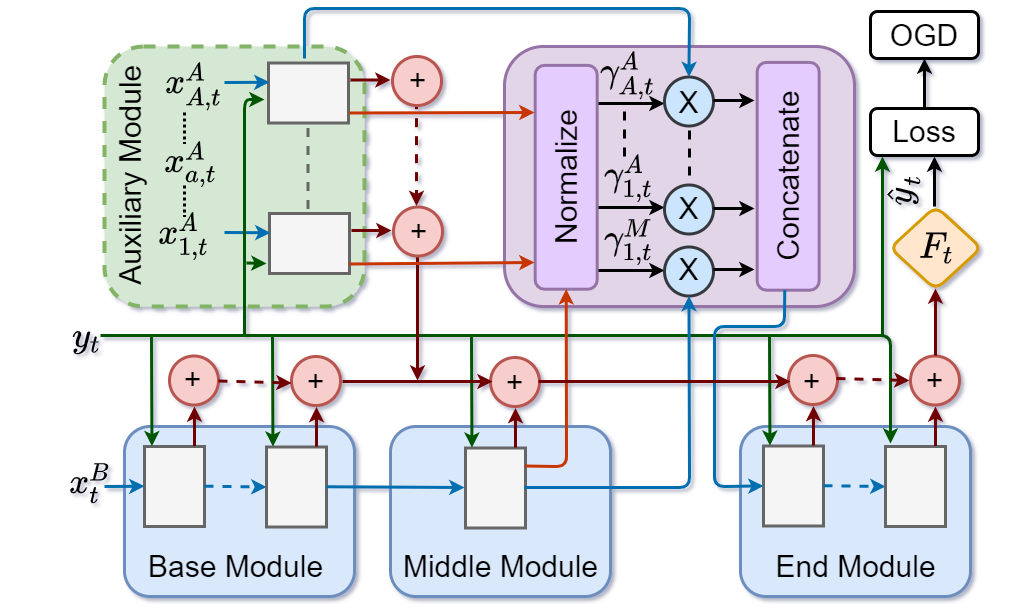}
  \caption{Detailed architecture of Aux-Net is presented here. The gray colored rectangular boxes represents a layer.}
  \label{fig3.1}
\end{subfigure}
\hspace{.6 mm}
\begin{subfigure}{.38\textwidth}
  \centering
  \includegraphics[width=1\linewidth]{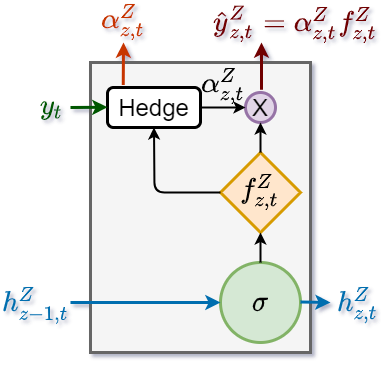}
  \caption{The functional diagram of a layer in figure \ref{fig3}(a) is shown here.}
  \label{fig3.2}
\end{subfigure}
\caption{The architecture of Auxiliary network (Aux-Net)}
\label{fig3}
\end{figure}


\section{Auxiliary Network (Aux-Net)}
\label{sec3}


\subsection{Problem Setting}
Let's denote the streaming classification data by $D = \{(x_1, y_1), ..., (x_t, y_t), ..., (x_T, y_T)\}$ where $x_t = \{x^B_{t}, x^{A_t}_t\}$ is the input at time instance $t$. The base features are denoted by $x^{B}_{t} = \{x^{B}_{1,t},...,x^{B}_{b,t},...,x^{B}_{B,t}\} $, where $B$ in superscript and subscript denotes the base features and total number of base features respectively and $x^{B}_{b,t}$ denotes the $b^{\rm{th}}$ base feature at time instance $t$.
The auxiliary features at any time instance $t$ is represented by $x^{A_t}_t = \{x^{A}_{j,t}\}_{\forall j \in A_t}$ where $A_t \subseteq \{1,...,a,...A\}$ is the subset of auxiliary features received at time instance $t$. The $A$ in superscript and subscript denotes the auxiliary features and total number of auxiliary features respectively and $x^{A}_{a,t}$ denotes the $a^{\rm{th}}$ auxiliary feature at time instance $t$.
The input $x_t \in \mathbb{R}^{d_t}$ where $d_t$ is the dimension of $x_t$ varying with time $t$ as shown in Figure \ref{fig1}.
The output $y_t \in \mathbb{R}^{\textit{C}}$ is the class label associated with $x_t$ where $\textit{C}$ is the total number of classes.
The Aux-Net learns a mapping $F:\mathbb{R}^{d_t} \rightarrow \mathbb{R}^{\textit{C}}$. The prediction of the model is given by $\hat{y_t} = F(x_t)$. The model trains in an online setting where at any time instance $t$, the input feature $x_t$ arrives, the model predicts an output $\hat{y_t}$, the actual output $y_t$ is revealed and an update is made to the model based on the loss $L(\hat{y_t}, y_t)$ incurred. Exhaustive list of all the notations is given in the supplementary.

\subsection{Architecture}
Consider a DNN with $S$ number of base layers, one middle layer, $A$ number of auxiliary layers and $E$ number of end layers. The base layers, middle layer, auxiliary layers and end layers constitute the base module, middle module, auxiliary module and end module respectively.
The base, middle and end modules are stacked sequentially and auxiliary module is placed in parallel to the base and middle module with a connection to the end module as shown in the Figure \ref{fig2}.
A softmax classifier is attached to each of the layer.
The detailed architecture of the model is presented in Figure \ref{fig3}.
The output of the Aux-Net model is given as the weighted combination of all the classifiers by the equation:

\vspace{-5 mm}
\begin{equation}
    F(x) = \sum\limits_{Z \in U} \sum\limits_{z=1}^{Z} \alpha^{Z}_{z}f^{Z}_{z}
    \label{eq3.1}
\end{equation}
where $U = \{S,M,E,A\}$ denotes all the modules, and $S,M,E,A$ in superscript and subscript denotes the module name and the total number of layers in the module respectively. The notation $f^Z_z$ and  $\alpha^Z_z$  represents the output of the classifier associated with the layer $z$ of the module $Z$ and weight of the classifier respectively.

The architecture of each layer is shown in Figure \ref{fig3}(b). Each layer is attached to a classifier $f$ parameterized by $\theta$ that gives an output as $ f^Z_z = \rm{softmax} (h^Z_z \theta^Z_z) $, where $h^Z_z$ is the hidden feature of the layer. Each layer is parameterized by $W$ and $c$ that takes the hidden feature of the previous layers as an input, and generates its hidden feature as $ h^Z_z = \sigma(W^{Z}_{z}h^{Z}_{z-1} + c^Z_z) $,
where $\sigma$ is the activation function and $\theta$, $W$ and $c$ are learnt using OGD approach.
A hedge block is used to compute $\alpha$ based on the loss incurred by the classifier.

Now, we describe the inputs to the different layers. The first base layer receives the complete $x^{B}_{t}$ as the input i.e. $h^Z_0 = x^{B}_{t}$. The subsequent base layers receive the hidden feature of its previous layer as its input. The middle layer receive the hidden feature of the last base layer as its input, i.e.  $h^M_0 = h^S_S$. The $a^{\rm th}$ auxiliary layer receives the $a^{\rm th}$ auxiliary feature as its input, i.e., $h^A_a = x^A_a$. All the end layers, except the first end layer receive its previous layer features as the input. The first end layer is special in terms of the input since the input to it needs to support agility arising from only a subset of auxiliary features being available at any time instance $t$. The input $h^E_{0,t}$ to the first end layer at time instance $t$ is a vector derived by concatenating weighted hidden features $h$ of the middle and the auxiliary layers corresponding to the currently available auxiliary inputs. It is therefore given by $
    h^E_{0,t} =\big[ \gamma^{M}_{1,t}h^{M}_{1,t},   \{  \gamma^{A}_{j,t} h^{A}_{j,t}\}_{\forall j \in A_t} \big]$,
where $\gamma$ is the importance of the layers connected to first end layer denoting the fraction of the connected layer's output passed as an input to the first end layer.

\subsection{Parameters Learning}
The learning of the model occurs in an online setting through the use of a loss function defined as:

\vspace{-5 mm}
\begin{equation}
    L(F(x),y) = \sum\limits_{Z \in U} \sum\limits_{z=1}^{Z} \alpha^{Z}_{z}L(f^{Z}_{z}(x), y)
    \label{eq3.5}
\end{equation}
where $L(f^{Z}_{z}(x), y)$ is the loss of the classifier associated with layer z of the module Z. On the basis of the loss incurred at each time step, the values of $\gamma, \theta, W, c, \alpha$ are updated as described next.

\textbf{Updating $\gamma$ :}
The highlight of Aux-Net is the update of $\gamma$ which allows for soft handling of the asynchronous availability of auxiliary features. It depends only on its classifiers weights and are calculated as follows:

\vspace{-5 mm}
\begin{equation}
\gamma^{P}_{p,t} =
\frac{\alpha^{P}_{p,t}}{\alpha^{M}_{1,t} + \sum_{j \in A_t} \alpha^{A}_{j,t}} \: \hspace{0.2cm} \text{for \: C1:} (P = M, \: p = 1) \: \rm or \: (P = A, \: p \in A_t)
\label{eq3.6}
\end{equation}

\textbf{Updating $\theta$ :}
The classifiers parameters $\theta$ is learned through OGD. The parameter $\theta^Z_z$ is associated with only one classifier and does not depend on the other classifiers. Therefore, its update will only be with respect to the loss of its own classifier.  After every time instance $t$, $\theta^Z_z$ of classifier $z$ of the module $Z$ is updated as:
\begin{equation}
    \theta^{Z}_{z,t+1} = \theta^{Z}_{z,t} - \eta\alpha^{Z}_{z,t}\Delta^Z_{\theta^{Z}_{z,t},z} \:  \hspace{.2cm}  \text{for \: C2:} (Z \in U', \: z \in \{1,...,Z\}) \: \rm or \: (Z = A, \: z \in A_t)
\label{eq3.7}
\end{equation}

$ \text{where}, \Delta^R_{\theta^{Z}_{z,t},r} = \frac{\partial L(f^{R}_{r}(x_t), y_t)}{\partial \theta^{Z}_{z,t}} $, $\eta$ is the learning rate of the parameters and $U' = \{S,M,E\}$.

\textbf{Learning $W$ and $c$ :}
The weights (W) and bias (c) of a layer are learned by back propagation on the final loss similar to OGD. But, since each layer is associated with a classifier unlike the traditional DNN where only last layer gives a prediction, the gradient descent is different. Here, the parameters of a layer depends on the loss of all its successive layers that directly or indirectly influence it. The following equation shows the weight update rule and the same is applicable for bias too.

\vspace{-5 mm}
\begin{equation}
    \begin{gathered}
    W^{A}_{a,t+1} = W^{A}_{a,t} - \eta \big[ \alpha_{a,t}^{A}\Delta^A_{W^{A}_{a,t},a} + \sum\limits_{e=1}^{E}\alpha_{e,t}^{E}\Delta^E_{W^{A}_{a,t},e}  \big] \\
    W^{Z}_{z,t+1} = W^{Z}_{z,t} - \eta \big[ \sum\limits_{j=z}^{Z}\alpha_{j,t}^{Z}\Delta^Z_{W^{Z}_{z,t},j} + \sum\limits_{Q=set} \sum\limits_{q=1}^{Q}\alpha_{q,t}^{Q}\Delta^Q_{W^{Z}_{z,t},q}  \big]\\
    \end{gathered}
    \label{eq3.8}
\end{equation}
where $set = \{M,E\}, \{E\}, \phi$ if $Z \in\ \{S\}, \{M\}, \{E\}$ respectively, and $z \in \{1,...,Z\}$.

\textbf{Learning $\alpha$ :}
We learn the value of $\alpha$ through hedge algorithm. Initially, the value of $\alpha$ is uniformly distributed i.e., $\alpha^Z_z = 1/L$, where $L$ is the total number of layers, $L=S+M+A+E$. The loss incurred by the classifier $z$ of module $Z$ at time instance $t$ is $L(f^Z_z(x_t), y_t)$ and its weight is $\alpha^{Z}_{z,t}$. The weights of the classifier are updated on the basis of its loss as:
\begin{equation}
\alpha^{Z}_{z,t+1} = \alpha^{Z}_{z,t} \beta^{L(f^Z_z(x_t), y_t)} \: \hspace{.8cm}  \text{for C2},
\label{eq3.9}
\end{equation}
where $\beta \in (0,1)$ is the discount rate parameter. There may come a situation where $\alpha^Z_z \rightarrow 0$. To avoid that since we don't want to neglect any layer, a smoothing parameter $\lambda$ is introduced where $\lambda \in (0,1)$. It ensures a minimum weight for each classifier by using the equation $\alpha^{Z}_{z,t+1} = \rm max(\alpha^{Z}_{z,t+1}, \lambda/L)
$.
The value of all $\alpha$ is then normalized such that 
$\sum\limits_{Z=U'} \sum\limits_{z=1}^{Z} \alpha^Z_{z,t+1} + \sum\limits_{j \in A_t} \alpha^A_{j,t+1} = 1$.


\subsection{Algorithm}

The Aux-Net is a test-then-train approach and since the number of auxiliary features are changing, the trained model learned at time step $t$ can't be used as it is for training or testing at time step $t+1$. We define a knowledge base $K$ which is updated after each time instance $t$.
We represent all the parameters of the knowledge base by  $'$ . The knowledge base $K$ at any time instance $t$ is given by
\begin{equation}
    K_t = \{W'_t, c'_t, \theta'_t, \alpha'_t\},
    \hspace{.2cm}
    \text{where} \hspace{.2cm} G_t = \{G^S_t, G^M_t, G^A_t, G^E_t\} \hspace{.2cm} \text{if} \hspace{.2cm} G \in \{W', c', \theta', \alpha'\}
    \label{eq3.11}
\end{equation}

\begin{wrapfigure}{r}{0.5\textwidth}
\centering
\includegraphics[scale=.28 ]{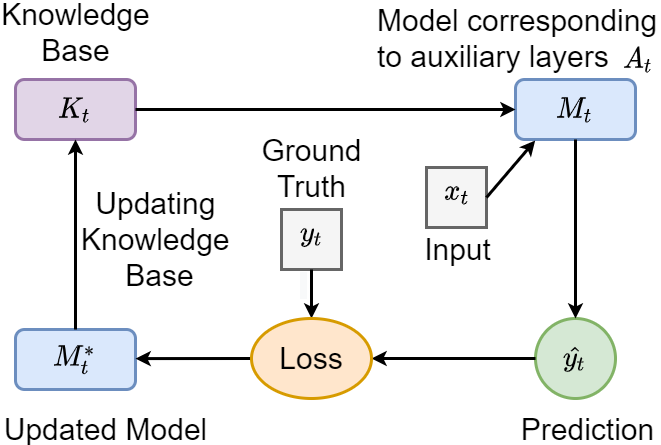}
\caption{Block diagram of Aux-Net algorithm. \vspace{-5 mm}}
\label{fig4}
\end{wrapfigure}

Before training or testing, the model needs to incorporate the incoming dynamic auxiliary features.
We define a model $M_t$ given by equation \ref{eq3.12}, that handles the asynchronous availability of auxiliary features ($A_t$) by introducing the variable $\gamma$. The model $M_t$ predicts an output $\hat{y_t}$, given $x_t$ and updates its parameter giving $M^{*}_t$ based on the loss incurred. Before moving to the next instance, we update the final parameters of $K_t$ based on $M^{*}_t$, giving knowledge base $K_{t+1}$. The block diagram of the algorithm is shown in Figure \ref{fig4} and algorithm is given in Algorithm \ref{Algo1}.

\textbf{Creating Model ($M_t$):} Based on the auxiliary features $A_t$ received at time step $t$ and knowledge base $K_t$, the model $M_t$ is created before prediction and training. The auxiliary layers corresponding to $A_t$ are kept active and all the other auxiliary layers are freeze. Freezing of layers means all the parameters associated with this layer will not be trained. In other terms, freezing means removing the layer from the model. Since, some of the auxiliary layers are removed, the value $\alpha$ of the model changes and a parameter $\gamma$ is introduced. The model $M_t$ is given by:
\begin{equation}
    M_{t} = M(W_{t}, c_{t}, \theta_{t}, \alpha_{t}, \gamma_{t})
\label{eq3.12}
\end{equation}

where $G_{t} = \{G'^S_{t}, G'^M_{t}, \{G'^{A}_{j,t}\}_{\forall j \in A_t}, G'^E_{t}\}$ if $G = \{W, c, \theta\}$,
$\alpha_t = \{\alpha^{Z}_{z,t}\}_{\forall \: \text{C2}}$ where
$\alpha^{Z}_{z,t} = $
$\alpha'^Z_{z,t}/ \bigg[\sum\limits_{Z=U'} \sum\limits_{z=1}^{Z} \alpha'^Z_{z,t} + \sum\limits_{j \in A_t} \alpha'^A_{j,t} \bigg]$, and
$\gamma_t = \{\gamma^{P}_{p,t}\}_{\forall \: \text{C1}}$,
$\gamma^{P}_{p,t} = $
$\alpha'^P_{p,t}/\bigg[ {\alpha'^{M}_{1,t} + \sum\limits_{j \in A_t} \alpha'^{A}_{j,t}} \bigg]$.

\begin{algorithm}[h]
  \SetAlgoLined
    \textbf{Inputs:} Base Module: $S$; Middle Module $M$; Auxiliary Module: $A$; End Module: $E$; Learning rate: $\eta$; Smoothing Parameter: $\lambda$; Discounting Parameter: $\beta$\;
    \textbf{Initialize:} A DNN with $L = S+M+A+E$ layers and attach classifiers to each layer as shown in Figure \ref{fig3.2}; $\alpha^Z_z = 1/L$ \: $\forall Z \in \{S,M,A,E\}, \: z \in \{1,...,Z\}$ ; $K_1$ using equation \ref{eq3.11}\;
    \SetAlgoNoLine
    \For{$t=1,...,T$}{
        Receive input feature $x_t$\;
        Create a list $A_t$ of the auxiliary features received in $x_t$\;
        Create the model $M_t$ based on $A_t$ using equation \ref{eq3.12}\;
        Predict $\hat{y}_t$ on $x_t$ using equation \ref{eq3.1} based on $M_t$\;
        Receive output label $y_t$\;
        Calculate the loss of the model $M_t$ based on $y_t$ and $\hat{y_t}$ using equation \ref{eq3.5}\;
        Update parameters of $M_t$ based on the loss incurred and get $M^{*}_{t}$ using \ref{eq3.13}\;
        Update $K_t$ based on $M^{*}_t$ to get $K_{t+1}$ using \ref{eq3.14}\;
        }
    \caption{Aux-Net algorithm}
    \label{Algo1}
\end{algorithm}
\textbf{Obtaining knowledge base $K_{t+1}$ for next instance:}
The parameters of the model $M_t$ are updated based on the loss incurred at time instance $t$. The updated model, represented by $M^*_t$ is given by:

\vspace{-5 mm}
\begin{equation}
    M^*_{t} = M(W^*_{t}, c^*_{t}, \theta^*_{t}, \alpha^*_{t})
    \label{eq3.13}
\end{equation}
where $W^*_{t}, c^*_{t}, \theta^*_{t}, \alpha^*_{t}$ are the parameters obtained by updating the parameters $W_{t}, c_{t}, \theta_{t}, \alpha_{t}$ of the model $M_t$ by using equation \ref{eq3.5}, \ref{eq3.7}, \ref{eq3.8}, and \ref{eq3.9}.
After training the model at time step $t$, we create the knowledge base $K_{t+1}$ before moving to the next iteration. All the parameters updated at time step $t$ and the parameters of the freezed layers ($A-A_t$) are collected. Then, $K_{t+1}$ is given by:

\vspace{-5 mm}
\begin{equation}
    K_{t+1} = \{W'_{t+1}, c'_{t+1}, \theta'_{t+1}, \alpha'_{t+1}\}\\
    \label{eq3.14}
\end{equation}
where $G'_{t+1} = \{G^*_t, \{G'^{A}_{j,t}\}_{\forall \: j \in A-A_t}\}$ if $G \in \{W, c, \theta\}$, and
$\alpha'_{t+1} = \{\alpha'^Z_{z,t+1}\}_{Z \in U, \: z = \{1,...,Z\}}$ where $\alpha'^Z_{z,t+1} = \alpha''^{Z}_{z,t}/ \bigg[ \sum\limits_{Z \in U} \sum\limits_{z=1}^{Z} \alpha''^{Z}_{z,t} \bigg]$ and
$\alpha''_{t+1} = \{\alpha^*_t, \{\alpha'^{A}_{j,t}\}_{\forall \: j \in A-A_t}\}$.



\section{Experimental Results}

\label{sec4}

We evaluate our model using the Italy power demand dataset \cite{dau2019ucr}. It has 1096 data instances with 24 input features. In all the studies, we retain the original order of the input features. To the best of our knowledge, there is no method that incorporates the intermittently available input data in an online setting.
Thus, we compare the Aux-Net model with the ODL model. We train both the models in a purely online setting where after each instance the model predicts and trains. 

\textbf{Architecture details} We fix the number of base layers ($S$) to be 5, the number of middle layer ($M$) is 1, and the number of end layers ($E$) is also 5 for Aux-Net. The number of auxiliary layers ($A$) are equal to the number of auxiliary features.
The number of layers for ODL is set as 11 ($S+E+M = 11$).
For both Aux-Net and ODL, we used the the ReLU activation function and the number of nodes in each layer was set as 50. The Adam optimizer ($\eta = 0.01$) was used for backpropagation. The smoothing rate and the discount rate was set as $\lambda$ = 0.2 and $\beta$ = 0.99 respectively. The cross-entropy loss was chosen as the loss function. The above settings are true for all the following experiments.

\textbf{Varying the probability $p$ of the availability of auxiliary inputs in Aux-Net} The first 12 input features of Italy power demand dataset are considered as the base features and remaining as the auxiliary features. The availability of each auxiliary feature at a given time instance is modeled as a uniform distribution with probability $p$. The same value of $p$ is used for all the auxiliary features but the availability of each is computed independently. The results of Aux-Net with varying values of $p$, ODL with all the 24 features, and ODL with only the 12 base features are presented in Table \ref{tab1}. We report the average of losses observed at all the time instances, and similarly the average accuracy observed across all the time instances. The cumulative average loss and accuracy curves are shown in Fig. \ref{fig5}. We study the performance of Aux-Net and comparison with ODL with the following aims:


\begin{figure}
    \centering

        \includegraphics[width=\textwidth]{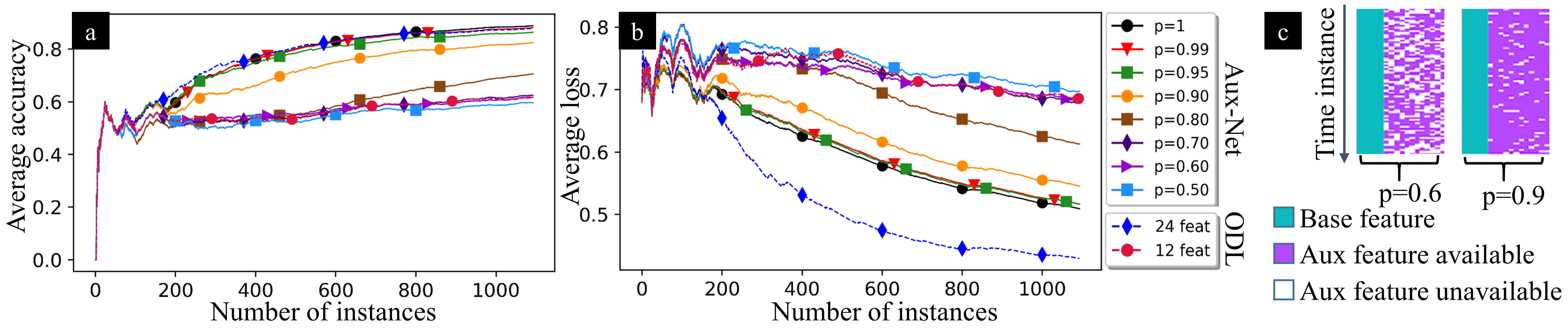}
    \caption{Cumulative average accuracy (a) and loss (b) for different values of probability $p$ of auxiliary inputs on Italy power demand dataset. ODL with 12 and 24 features is included for baseline. Snippet of data availability for $p=0.6$ and $p=0.9$ are shown in (c), analogous to Figure \ref{fig1}.}
    \label{fig5}
\end{figure}

$\bullet$ Sensitivity of Aux-Net to $p$ and its performance: The average accuracy and loss for all the time instances in the dataset shows monotonic trend as a function of $p$, as noted in Table \ref{tab1}. This shows that Aux-Net is sensitive to the availability of the auxiliary inputs, as expected. Yet, the performance of Aux-Net degrades gracefully as $p$ reduces. Moreover, Aux-Net still performs better compared to ODL with 12 features when $p<1$ (as ODL can not work with inconsistent features). Further, the best case performance of Aux-Net when $p=1.00$, is comparable to the scenario of ODL with 24 features. This means that even though the knowledge base of Aux-Net supports for $2^{12}$ knowledge models, only the knowledge model with largest dimensionality is invoked and trained. In this case loss of Aux-Net is poorer than ODL, but the accuracy is better. In case of $p=0.5$ which means no consistency in either availability or unavailability of the auxiliary inputs the observed poorer performance of Aux-Net in comparison to ODL is only marginal, indicating robustness of Aux-Net to the extremely challenging scenario and its graceful degradation.
\begin{wraptable}{r}{6.8cm}
\vspace{-5 mm}
  \caption{Average accuracy and loss of Aux-Net (for different values of probability($p$) of availability of auxiliary features) and ODL (with different number of input features(feat)) in Italy Power Demand dataset. \textit{ODL} is shown in italics.}
  \label{tab1}
  \centering

  \begin{tabular}{lcc}
    \toprule
    Model      & Accuracy  & Loss \\
    \midrule
    \textit{ODL(24 feat)}  &  \textit{0.8783} &  \textit{0.4297} \\
    \cellcolor[gray]{0.9}Aux-Net($p = 1.00$)  &  \cellcolor[gray]{0.9}{0.8884} & \cellcolor[gray]{0.9}0.5093 \\
    \cellcolor[gray]{0.9}Aux-Net($p = 0.99$)  &  \cellcolor[gray]{0.9}0.8811 &  \cellcolor[gray]{0.9}0.5165 \\
    \cellcolor[gray]{0.9}Aux-Net($p = 0.95$)  &  \cellcolor[gray]{0.9}0.8637 &  \cellcolor[gray]{0.9}0.5168 \\
    \cellcolor[gray]{0.9}Aux-Net($p = 0.90$)  &  \cellcolor[gray]{0.9}0.8243 & \cellcolor[gray]{0.9}0.5456 \\
    \cellcolor[gray]{0.9}Aux-Net($p = 0.80$)  &  \cellcolor[gray]{0.9}0.7054 &  \cellcolor[gray]{0.9}0.6130 \\
    \cellcolor[gray]{0.9}Aux-Net($p = 0.70$)  &  \cellcolor[gray]{0.9}0.6240 &  \cellcolor[gray]{0.9}0.6788 \\
    \cellcolor[gray]{0.9}Aux-Net($p = 0.60$)  &  \cellcolor[gray]{0.9}0.6167 &  \cellcolor[gray]{0.9}0.6831 \\
    \textit{{ODL(12 feat)}}  &  \textit{{0.6139}} &  \textit{{0.6868}} \\
    \cellcolor[gray]{0.9}Aux-Net($p = 0.50$)  &  \cellcolor[gray]{0.9}{0.5956} &  \cellcolor[gray]{0.9}0.6975 \\\bottomrule
  \end{tabular}
  \vspace{-5 mm}
\end{wraptable}
$\bullet$ Agile adaptation of Aux-Net: The demand on agility significantly enhances as $p$ reduces. For example, for $p=0.6$ in Figure \ref{fig5}(c), not only a different knowledge model needs to be invoked at every instance but also the same knowledge model may not be invoked in next many time instances. The situation is easier when $p=0.9$ even though there are many time instances when a different knowledge model is invoked. Nonetheless, Aux-Net remains stable in either case and adapts to the agility needs in an efficient manner, indicated in accuracy and loss plots in Figure \ref{fig5}(a,b). Indeed, the accuracy is better and the loss decreases faster over time for $p=0.9$. Nonetheless, when $p=0.6$, the accuracy and loss curves closely follow ODL with 12 features, indicating that even though new knowledge models are being dynamically invoked every single instance, the performance of Aux-Net does not deteriorate in comparison to ODL and Aux-Net is indeed able to maintain agility over time, contributing to reduced loss and improved accuracy as time passes.

$\bullet$ Decreased loss and improved accuracy over time: We note that for the situation of 12 auxiliary inputs, support for $2^{12}$ knowledge models, and invocation of each knowledge model multiple times is needed to study the aspects such as convergence of knowledge base over time. Yet, the decreasing loss in Figure \ref{fig5}(b) is a positive indicator of performance improvement over time and possible convergence.

\begin{figure}[t]
\centering
\includegraphics[scale=.37]{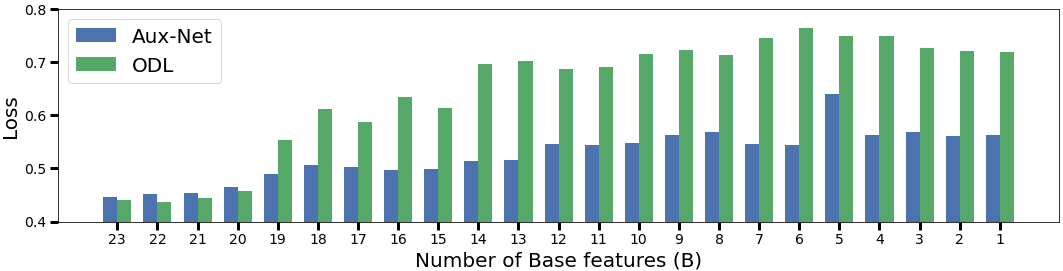}
\caption{Loss of Aux-Net as function of the number of base features $B$ and ODL (trained using $B$ number of features) in Italy power demand dataset.
The probability of availability of the $(24-B)$ auxiliary inputs is fixed at $p=0.9$.
Lower loss indicates better learning.}
\label{fig6}
\end{figure}

\textbf{Varying number of base features} In this experiment, we fix $p$ as 0.9, but vary the number of base features (B) from 1-23. The number of auxiliary features (A) are consequently (24-B). The first $B$ features in the dataset are used as base features in Aux-Net and the only features in ODL. The average loss of Aux-Net and ODL are compared in Fig. \ref{fig6} as a function of $B$. We observe the following:

$\bullet$ Extreme scalability: As expected, the performances of both Aux-Net and ODL deteriorate as $B$ reduces. Nonetheless, the loss of Aux-Net is significantly smaller than ODL in the challenging scenarios when more than 4 inputs are inconsistently available. This clearly indicates that Aux-Net is able to leverage the auxiliary inputs for better learning even if they are inconsistently available.
Especially, the extremely challenging scenarios ($B$ = 1 for example) demonstrate that Aux-Net is indeed able to step up to the need of supporting several knowledge models of varying inputs and dimensionalities and provide better performance than the minimalist approach.

$\bullet$ Poorer performance than ODL when $B \in [20,23]$: During initialization, Aux-Net assigns the same weights ($\alpha$) to each classifier. However, the classifier corresponding to an auxiliary feature will be lossier as compared to the classifier of middle layer that uses base features. As time progresses, the value of $\alpha$ for each layer gets customized to suit its contribution towards accurate classification. Often, it means that $\alpha$ of auxiliary layer reduces in the first few time instances, indicating that Aux-Net has learnt that its inconsistent availability may cause increased loss if $\alpha$ corresponding to it is high.

\section{Discussion and conclusion}\label{sec5}

\textbf{Scalability and knowledge entity:} Aux-Net supports scalability for the situations ranging from no auxiliary input being available to all auxiliary inputs being available. Aux-Net incorporates knowledge models corresponding to all possible combinations of auxiliary inputs within a single knowledge base. The architectural support in Aux-Net for auxiliary inputs in the form of dedicated parallel layers is a critical feature for scalability. At the same time, being able to update the pertinent knowledge models selectively and reflect the new knowledge back into the main knowledge base (see Fig. \ref{fig4}) ensures that a single knowledge base needs to be maintained as opposed to the resource-heavy maximalist approach. Further more, online learning in the current framework dispenses away the need of offline storage of data. Nonetheless, in the future application-specific framework, maintaining a stash of offline data may provide an advantage and exploring that is a possibility.

\textbf{Agility and stability:} Agility in Aux-Net is characterized by its ability to dynamically invoke the relevant knowledge model without making the network unstable or unadaptive. A key factor that supports dynamic stability and agility is the importance parameter $\gamma$, which automatically adjusts the contributions of base inputs (through the middle layer) and the currently available auxiliary inputs so that neither the new auxiliary features introduce inordinate instability, nor are they suppressed.

It is of interest to investigate the convergence of Aux-Net, which could not be investigated rigorously in this study though indicators of convergence were observed (results in the supplementary). Since there was no possibility so far for dealing with intermittently available inputs, we found that there is a dearth of suitable benchmark datasets and applications, which allow us to empirically assess aspects such as convergence. However, we deem that more elaborate studies are needed on suitably designed datasets and investigation of rigorous theoretical proofs of convergence will be significantly useful.

So far, we have demonstrated scalability, agility, and stability of Aux-Net and its ability to deal with intermittently available inputs in a completely online manner. This, in our observation is not only the first such architecture, it is also a first demonstration of results on intermittently available input features. Having set a new paradigm, we hope that new datasets, new frameworks, new applications, and more extensive studies are developed in the near future to exploit the possibility of learning in extremely dynamic and uncertain scenarios. We hope that advanced artificial intelligence for dynamic complex environments will soon emerge. We will work on providing further conceptual groundwork to Aux-Net, identifying or creating new benchmark datasets, extending Aux-Net to perform tasks such as prediction and detection or deal with more variety of inputs, such as images, adapt it to use convolutional kernels, and working with asynchronous multi-modal inputs in the future.

Machine learning community has been afflicted by rigid architectures for long even though activities in extreme learning, neuro-evolution, and incremental or online learning have been explored to ease the problem of architectural rigidity. Yet, the dream to perform advanced artificial intelligence in a highly dynamic, situation adaptive manner for efficient operation in real-world dynamic complex environments is far from accomplished. Even the most advanced AI agents, such as autonomous cars considers rigid architecture as the amount and type of data availability changes. We are all waiting for a truly agile, scalable, self-adapting non-rigid artificial intelligence approach that redefines how intelligent machines deal with varying amount and types of input features and ad hoc environment.

Aux-Net is a baby step towards the above mentioned dream. For the first time (in our knowledge) we have showcased that an architecture can be online, scalable and agile in nature. In the future, scalable and agile machine learning will bring the next wave of research and development activities, which can address the pressing needs of advanced machine learning in complex dynamic environment. To support the initial activity on this new direction, we will release the Aux-Net source code and its development platform for the benefit of the further research and development activities.

Aux-Net in its current form has its limitations. It needs to be developed for online image based classification similar to the current state-of-the-art deep learning architecture. It has been currently built on keras with tensorflow backend, which currently lack support for Aux-Net type online scalable and agile learning. We will soon release the basic functionality libraries which can support the scalable, agile and online learning. We invite researchers for developing suitable benchmarking datasets from dynamic complex environment with scalable and agile machine learning needs as well as contribute to better implementation and other AI tasks.

\bibliographystyle{unsrtnat}
\bibliography{mybib}   

\end{document}